\documentclass[letterpaper, 10 pt, conference]{ieeeconf}  

\IEEEoverridecommandlockouts                              

\usepackage{amsmath,amssymb,amsfonts} 
\usepackage{graphicx} 
\usepackage{booktabs} \usepackage{xcolor}
\usepackage{cite}
\usepackage{url}
\usepackage{multirow}
\newcommand{\AVAR}{\operatorname{AV@R}}  
\newcommand{\VAR}{\operatorname{V@R}}  

\title{\LARGE \bf
SemSafe-3DGS: Semantic Risk-Aware Active Navigation in Uncertain 3D Gaussian Splatting Maps
}

\author{Amirhossein Mollaei Khass, Athanasios Cosse, Nader Motee
\thanks{
This work was supported in part by the ONR under grants numbers  N00014-23-1-2779 and N00014-26-1-2246. \newline
A.M. Khass, A. Cosse, and N. Motee are with the Department of Mechanical Engineering and Mechanics, Lehigh University, Bethlehem, PA, 18015, USA. {\tt\small \{ammb23,asc425,vkp219,motee\}@lehigh.edu}.\endgraf
}
}

\begin{document}

\maketitle
\thispagestyle{empty}
\pagestyle{empty}

\begin{abstract}
Autonomous robots operating in partially observed environments must navigate safely while acquiring observations that improve future planning. Existing safety formulations generally reason primarily about geometry. Consequently, geometrically similar scene elements may induce comparable control responses despite having different semantic consequences. We present a semantic risk aware safe-active perception framework for navigation in attributed 3D Gaussian maps. Semantic attributes modulate an Average Value-at-Risk collision clearance model through class dependent risk weights, allowing safety-critical Gaussian primitives to receive greater influence in the composite barrier. The resulting weighted clearances are aggregated into a control barrier function, while a trajectory-relevant active perception barrier promotes observations that reduce geometric map uncertainty along the robot's anticipated motion. Both objectives are integrated in a unified CBF-QP that enforces semantic risk-aware collision avoidance as a hard constraint while relaxing information acquisition when it conflicts with safety or task progress. Experiments demonstrate efficient safety constraint, improved navigation through active perception, semantic dependent trajectory adaptation, and real-robot execution under Ackermann dynamics.
\end{abstract}
\section{Introduction} Autonomous robots navigating unknown environments must make decisions from uncertain maps.
An effective policy must jointly determine the robot’s motion and sensing direction to construct a map that supports collision avoidance and task execution.
This introduces an inherent conflict, as informative observations often require directing the sensor toward occluded or poorly modeled regions where the corresponding safety margins are least reliable.

3D Gaussian Splatting (3DGS) provides a reliable representation for robotic mapping because it combines explicit spatial primitives with differentiable RGB-D rendering~\cite{kerbl20233d}.
Recent methods augment Gaussian maps with semantic labels, language features, and task-relevant attributes~\cite{qin2024langsplat, li2024gs3lam}. However, these attributes are primarily used for recognition and scene understanding rather than safe navigation.
Existing 3DGS based navigation methods, however, primarily reason about geometric structure~\cite{chen2025splat, chen2025control,khass2026conflict,khass2026multi}. Consequently, semantically distinct entities may derive comparable control responses when their geometric configurations are similar.

We propose a semantic risk aware active perception framework for online 3DGS navigation. Semantic attributes modulate the relative collision risk associated with each Gaussian, while the active perception objective focuses on reducing uncertainty near the robot's future trajectory.
A smooth composite safety CBF enforces safety, and perception CBF promotes informative acquisition. The two constraints are integrated in a unified quadratic program that preserves safety as a hard requirement and relaxes perception when conflict occur.

Our contributions are: 
(i) A semantic risk aware control policy that integrates geometric
uncertainty with attributed 3D Gaussian maps. (ii) A smooth composite safety CBF that captures collision risk from all Gaussian obstacles through a single differentiable constraint. (iii) A unified CBF-QP that couples semantic risk-aware safety with trajectory relevant active perception.

\begin{figure}[t]
    \centering
    \includegraphics[
        width=\columnwidth,height=0.18\textheight,
        trim={0.25cm 0.30cm 0.20cm 0.15cm},
        clip
    ]{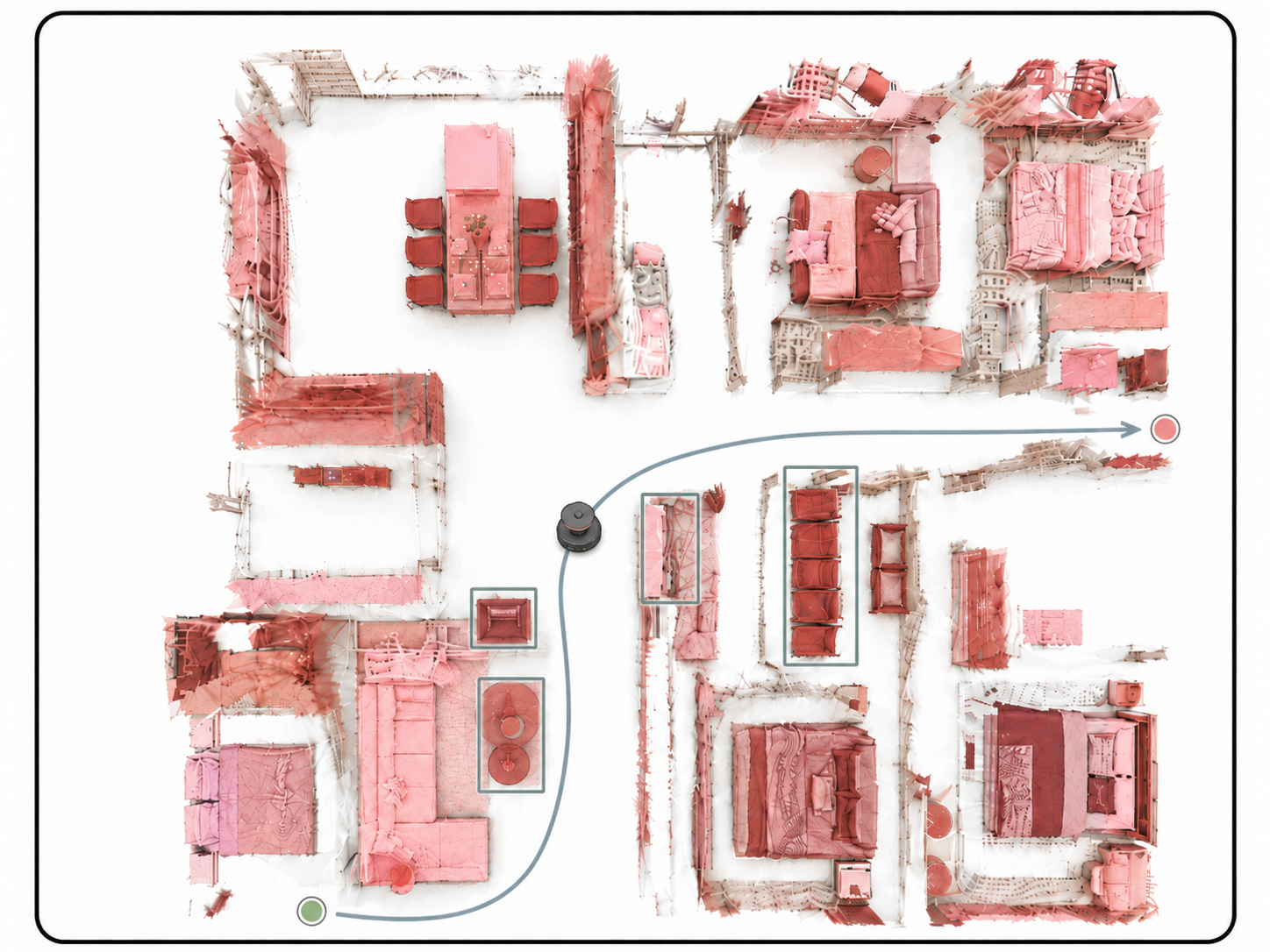}
    \vspace{-6.5mm}
\caption{
Semantic 3DGS map with a safe trajectory and nearby semantically labeled objects that influence collision risk.
}
\label{fig:semantic_trajectory}
    \vspace{-2mm}
\end{figure}

\section{Problem Formulation} We consider a robot navigating toward a goal in an environment with limited prior knowledge.
At each control step, the robot receives an onboard RGB-D observation and updates an online Gaussian map.
The objective is to preserve safety while moving toward the goal and collect observations that improve the map in regions relevant to future motion.
The robot is modeled as the control-affine system
\begin{equation} 
\dot{x}=f(x)+g(x)u, 
\label{eq:dynamics}
\end{equation}
where \(x\in\mathcal X\subset\mathbb R^n\) is the state and \(u\in\mathcal U\subset\mathbb R^m\) is the control input.
Let \(\pi(x)\in\mathbb R^3\) and \(\eta(x)\in\mathbb S^2\) denote the robot position and camera viewing direction, respectively.
The environment is represented by an attributed 3DGS map
\begin{equation} 
\mathcal G_k=\{g_i^k\}_{i=1}^{M_k},
\qquad 
g_i=(\mu_i,R_i,S_i,o_i,c_i,\xi_i),
\label{eq:gaussian_map}
\end{equation}
where \(\mu_i\), \(R_i\), \(S_i\), \(o_i\), and \(c_i\) denote the Gaussian position, orientation, scale, opacity, and appearance.
Its covariance is $\Sigma_i=R_iS_iS_i^\top R_i^\top, $
and \(\xi_i\) denotes its semantic attribute, such as a class distribution and a semantic feature with an associated confidence.

\textbf{Problem Statement.}
The objective is to learn a control policy that generates safe and informative robot actions by enforcing semantic-risk-aware collision avoidance as a hard constraint while promoting trajectory-relevant information acquisition through a soft perception objective.

\section{Semantic-Risk-Aware Safe-Informative Control} \subsection{Semantic-Risk-Aware Safety Barrier}
The safety objective is to prevent the robot from approaching uncertain or semantically hazardous regions represented by the current 3DGS map. 
Let \(d_i(\pi)\) denote the uncertain distance between the robot position and Gaussian \(g_i\).
We quantify its lower-tail clearance using Average Value-at-Risk: \begin{equation}
    \mathrm{\AVAR}_{\varepsilon}(d_i^x)
    :=
    \mathbb{E}\!\left[
        d_i^x \;\middle|\; d_i^x < \mathrm{\VAR}_{\varepsilon}(d_i^x)
    \right]
\qquad
\varepsilon\in(0,1),
\label{eq:avar}
\end{equation}
where \(\operatorname{VaR}_{q}(d_i)\) is the \(q\)-quantile of the distance distribution.
The AV@R therefore measures the average clearance within the lower tail of the distance distribution and provides a conservative representation of geometric uncertainty.
We define a positive semantic risk weight
\begin{equation}
    \kappa(\xi_i) > 0,
\end{equation}
which maps the semantic attribute and its confidence to the relative
collision risk associated with Gaussian $g_i$. More safety-critical or
semantically uncertain objects are assigned smaller values of
$\kappa(\xi_i)$, reducing their positive geometric clearance and increasing
their influence on the composite safety barrier. The semantic risk aware
clearance is then
\begin{equation}
    \rho_i^{\mathrm{sem}}(\pi)
    =
    \kappa(\xi_i)
    \left(
    \operatorname{AV@R}_{\varepsilon}(d_i(\pi))
    -r_{\mathrm{rob}}
    \right).
    \label{eq:clearance}
\end{equation}
where \(r_{\mathrm{rob}}\) is the inflated robot radius. A nonnegative value indicates that the lower-tail distance remains outside the collision margins~\cite{khass2025active}.

Safety with respect to the complete Gaussian map requires
\begin{equation}
\min_{g_i\in\mathcal G_k} \rho_i^{\mathrm{sem}}(\pi) \ge 0. \end{equation} 
Because the pointwise minimum is nonsmooth, we approximate it by the composite safety barrier
\begin{equation}
h_s(\pi) =
-\frac{1}{\beta_s} \log \left( \sum_{g_i\in\mathcal G_k} \exp \left[-\beta_s\rho_i^{\mathrm{sem}}(\pi)\right] \right),
\qquad 
\beta_s>0. 
\label{eq:safety_barrier}
\end{equation} 
For \(M_k\) Gaussian primitives~\cite{khass2026conflict,molnar2023composing}, 
\begin{equation}
\min_i\rho_i^{\mathrm{sem}} - \frac{\log M_k}{\beta_s}
\le 
h_s 
\le 
\min_i\rho_i^{\mathrm{sem}}.
\label{eq:softmin_bound}
\end{equation} 
Thus, \(h_s(\pi)\ge0\) is a smooth sufficient condition for maintaining nonnegative semantic risk of collision. For the dynamics in~\eqref{eq:dynamics}, the corresponding safety CBF condition is \begin{equation}
L_fh_s(x)+L_gh_s(x)u \ge -\gamma_s\alpha_s(h_s(x)),
\label{eq:safety_cbf}
\end{equation}
where \(\alpha_s\) is an extended class-\(\mathcal K\) function and \(\gamma_s>0\). 
The soft-min structure also provides smooth attention over the 3DGS map: \begin{equation}
\nabla_{\pi}h_s
= 
\sum_{i=1}^{M_k} \omega_i \nabla_{\pi}\rho_i^{\mathrm{sem}},
\qquad 
\omega_i = \frac{ e^{-\beta_s\rho_i^{\mathrm{sem}}} }{ \sum_j e^{-\beta_s\rho_j^{\mathrm{sem}}} }. 
\label{eq:safety_attention}
\end{equation} 
Consequently, the control response is dominated by Gaussian primitives
with the smallest semantic-weighted clearance. By assigning smaller
positive values of $\kappa(\xi_i)$ to higher-risk semantic classes,
such primitives receive greater weight in the composite barrier and can
influence the control action earlier than geometrically similar benign
objects.

\subsection{Trajectory-Relevant Active-Perception Barrier} 
Safety alone does not guarantee that the robot acquires observations that improve future planning. We therefore retain a trajectory-relevant geometric active-perception objective that evaluates information gain only within regions associated with the robot’s anticipated motion.
Let $\mathcal P_k= \{\bar p_0,\bar p_1,\ldots,\bar p_H\} $,
denote a nominal trajectory by \(u_k^{\mathrm{ref}}\), where \(\bar p_0=\pi(x_k)\). 
In each waypoint, we define
\begin{equation}
r_\tau = r_{\min} + \beta_1 \exp \left( -\beta_2 \min_{g_i\in\mathcal G_k} \rho_i^{\mathrm{sem}}(\bar p_\tau) \right),
\label{eq:mask_radius} 
\end{equation}
The trajectory relevant map region is
\begin{equation} \Omega_k = \bigcup_{\tau=0}^{H} \mathcal B(\bar p_\tau,r_\tau), \qquad \mathcal G_k|_{\Omega} = \{g_i\in\mathcal G_k\mid\mu_i\in\Omega_k\}. 
\label{eq:perception_mask} 
\end{equation}
This restricts information evaluation to map regions likely to affect future navigation. 

Let \(w_{\Omega}\) denote the parameters of \(\mathcal G_k|_{\Omega}\) and $Y$ be the RGB-D observation predicted from sensing pose \((\pi,\eta)\), the observed Fisher information is approximated by
\begin{equation}
\mathcal H((\pi,\eta)) = -\nabla_{w_{\Omega}}^2 \log p(Y\mid (\pi,\eta),w_{\Omega}) \big|_{w_{\Omega}=w_{\Omega}^{\star}}. 
\end{equation}
Let \(\mathcal H_{\mathrm{prior}}\succ0\) denote the regularized information accumulated from previous observations.
We define the trajectory relevant information utility as
\begin{equation} \mathcal I((\pi,\eta)) = \operatorname{tr} \left( {\mathcal H}((\pi,\eta)) \mathcal H_{\mathrm{prior}}^{-1} \right). \label{eq:information_utility} \end{equation} 
The information score increases when the candidate sensing pose is expected to reduce uncertainty in the Gaussian parameters associated with the robot’s nominal future trajectory~\cite{jiang2024fisherrf,khass2025active}.
The active perception barrier is therefore defined as
\begin{equation}
h_p(\pi,\eta) = \mathcal I(\pi,\eta)-\mathcal I_c, 
\label{eq:perception_barrier}
\end{equation}
where \(\mathcal I_c>0\) is a desired information threshold. Its CBF condition is
\begin{equation}
L_fh_p(x)+L_gh_p(x)u \ge -\gamma_p\alpha_p(h_p(x))-\delta, 
\qquad 
\delta\ge0.
\label{eq:perception_cbf}
\end{equation} 
The slack variable allows information acquisition to be relaxed when it conflicts with collision avoidance, control limits, or nominal task progress.

\begin{figure*}[t]
    \centering

    \begin{minipage}[c][4.25cm][c]{0.675\textwidth}
        \centering
        \includegraphics[
            width=\linewidth,
            trim={0.4cm 1.9cm 0.2cm 2.38cm},
            clip
        ]{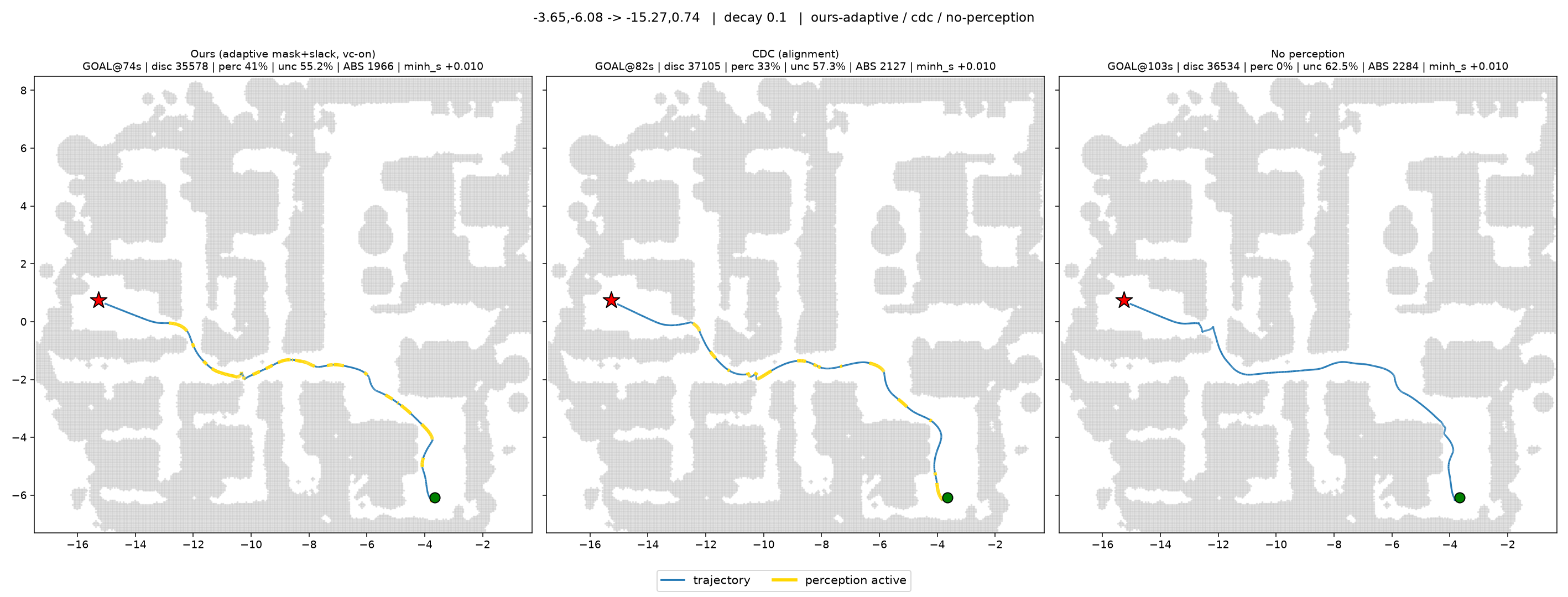}
    \end{minipage}
    \hfill
    \begin{minipage}[c][4.25cm][c]{0.305\textwidth}
        \centering
        \footnotesize
        \setlength{\tabcolsep}{1.0pt}
        \renewcommand{\arraystretch}{1.12}

        \resizebox{\linewidth}{!}{%
        \begin{tabular}{@{}lccc@{}}
            \toprule
            & \textbf{Ours} & CAAP & No perc. \\
            & \textbf{} &  \textbf{} & \\
            \midrule
            Goal time [s]
                & \textbf{74}
                & 82
                & 103 \\

            Path length [m]
                & 16.6
                & 16.9
                & \textbf{16.4} \\

            Mean speed [m/s]
                & \textbf{0.212}
                & 0.192
                & 0.143 \\

            \midrule
            Mean $h_s$
                & $\mathbf{+0.724}$
                & $+0.680$
                & $+0.561$ \\

            Mean ${h}_p$
                & $\mathbf{+0.005}$
                & $-0.275$
                & --- \\
            \bottomrule
        \end{tabular}%
        }
    \end{minipage}

    \vspace{-1.5mm}
\caption{
Safe-active navigation comparison in InteriorGS: proposed method (left), CAAP~\cite{khass2026conflict} (middle), and safety-only navigation (right). Yellow trajectory segments indicate active perception. Our method achieves the shortest goal time and highest mean speed while maintaining positive mean safety and perception barriers.
}
\label{fig:perception_comparison}
    \vspace{-2mm}
\end{figure*}

\subsection{Unified Semantic Safe-Active CBF-QP}
The next safe-informative control policy is obtained from \begin{equation}
\begin{aligned} 
(u_k^\star,\delta_k^\star) 
= 
\arg\min_{u,\delta} 
\quad&
\frac{1}{2} \|u-u_k^{\mathrm{ref}}\|_2^2 + \frac{\lambda_\delta}{2}\delta^2
\\ 
\mathrm{s.t.}
\quad
& 
L_fh_s+L_gh_su \ge -\gamma_s\alpha_s(h_s), 
\\
& 
L_fh_p+L_gh_pu \ge -\gamma_p\alpha_p(h_p)-\delta,
\\ 
&
u_{\min}\preceq u\preceq u_{\max},
\qquad
\delta\ge0,
\end{aligned} 
\label{eq:unified_qp}
\end{equation} 
where \(\lambda_\delta>0\) controls the cost of relaxing the perception objective. The first constraint enforces semantic-risk-aware collision avoidance and is never relaxed. The second constraint promotes geometric information acquisition near the robot's anticipated trajectory but permits controlled violation through \(\delta\). The objective minimally modifies the nominal task policy while penalizing the loss of perception performance. The resulting controller therefore prioritizes safety around geometrically uncertain or semantically hazardous regions while continuing to improve the map whenever informative sensing is compatible with safe task execution.


\begin{table}[t]
\centering
\scriptsize
\setlength{\tabcolsep}{4.5pt}
\renewcommand{\arraystretch}{0.8}

\begin{tabular}{llcccc}
\toprule
\textbf{Method} &
\textbf{Scene} &
\shortstack{\textbf{Safety}\\\textbf{[\%]} $\uparrow$} &
\shortstack{\textbf{Time}\\\textbf{[ms]} $\downarrow$} &
\shortstack{\textbf{Min. Dist.}\\\textbf{[m]} $\uparrow$} &
\shortstack{\textbf{Control}\\\textbf{Dev.} $\downarrow$} \\
\midrule

\multirow{4}{*}{SAFER-Splat}
& Stonehenge  & 100 & 30.9 & 0.086 & 0.069 \\
& Statues     & 99  & 45.0 & 0.141 & 0.028 \\
& Flightgate  & 99  & 60.3 & 0.086 & 0.062 \\
& Adirondacks & 100 & 186.0 & 0.079 & 0.046 \\
\midrule

\multirow{4}{*}{\shortstack[l]{Ours\\$c=0.90$}}
& Stonehenge  & 100 & 2.64 & 0.120 & 0.105 \\
& Statues     & 99  & 2.68 & 0.169 & 0.104 \\
& Flightgate  & 98  & 2.72 & 0.150 & 0.098 \\
& Adirondacks & 98  & 2.79 & 0.111 & 0.098 \\
\midrule

\multirow{4}{*}{\shortstack[l]{Ours\\$c=0.75$}}
& Stonehenge  & 100 & 1.63 & 0.106 & 0.070 \\
& Statues     & 99  & 1.49 & 0.158 & 0.024 \\
& Flightgate  & 99  & 1.50 & 0.143 & 0.064 \\
& Adirondacks & 99  & 1.84 & 0.075 & 0.044 \\
\bottomrule
\end{tabular}
\caption{Safety-barrier comparison across four 3DGS scenes.}
    \vspace{-6.5mm}
\label{tab:safety_comparison}
\end{table}


\section{Experiments}
\label{sec:experiments}

We evaluate the proposed framework through different complementary studies: 
(i) computational and performance of the composite safety barrier, 
(ii) the effect of trajectory-relevant active perception on closed-loop navigation, and 
(iii) real-robot deployment under Ackermann dynamics.

\subsection{Safety Barrier Evaluation}
We first isolate the safety component using double-integrator dynamics and compare it with SAFER-Splat~\cite{chen2025control}, which constructs collision constraints from the distance between the robot and individual Gaussian ellipsoids. For a controlled comparison, all methods use the same PD nominal controller, dynamics, and 3DGS maps trained with Splatfacto~\cite{Xu2024SplatfactoWAN}: Stonehenge ($100$K Gaussians), Statues ($200$K), Flightgate ($300$K), and Adirondacks ($500$K). As reported in Table~\ref{tab:safety_comparison}, SAFER-Splat achieves an average safety rate of $99.5\%$, compared with $98.75\%$ and $99.25\%$ for our barrier with $c= 1- \epsilon \in \{0.75, 0.90\}$, respectively. A more conservative risk estimate enlarges the effective obstacle region and can slightly reduce the success rate in cluttered scenes by restricting admissible motions. In contrast, the average online computation decreases from $80.55$\,ms to $2.71$\,ms and $1.62$\,ms, corresponding to approximately $30\times$ and $50\times$ speedups. The proposed barrier also maintains larger average minimum clearances for both settings, demonstrating that the smooth map-level aggregation substantially reduces computation without compromising collision avoidance.

\begin{figure*}[t]
    \centering
    \includegraphics[
        width=0.96\textwidth,height=0.22\textheight,
        trim={4.1cm 6.5cm 4.1cm 3.1cm},
        clip
    ]{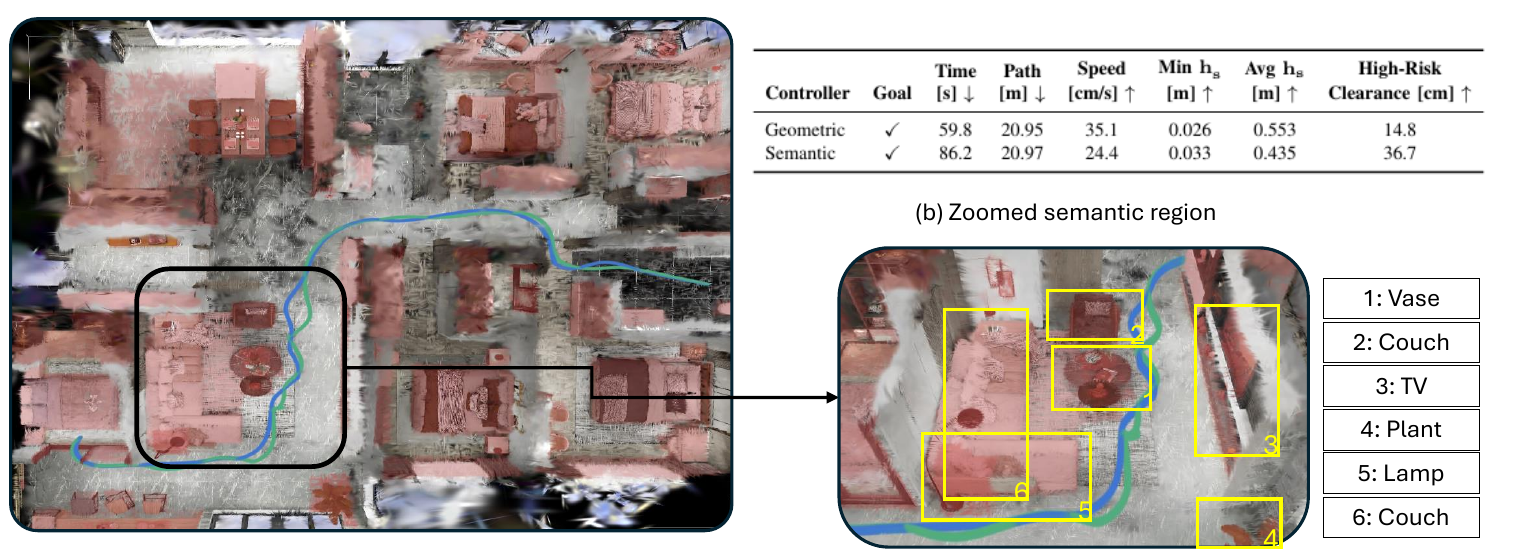}
\caption{
Semantic-risk-aware navigation in an attributed 3DGS map.
(a) Executed trajectories with geometric safety + perception (blue) and
semantic-aware safety + perception (green).
(b) Enlarged region showing semantic object annotations used to modulate
class-dependent collision-risk weights.
}
    \label{fig:semantic_risk_comparison}
\end{figure*}


\subsection{Trajectory-Relevant Active Perception}
We next compare the complete controller with CAAP~\cite{khass2026conflict} and a safety-only baseline in the InteriorGS~\cite{InteriorGS2025} environment using unicycle dynamics. The safety-only controller removes the perception constraint and follow the nominal trajectory safety without any information seeking, while CAAP promotes alignment with the expected information gain ascend direction. All methods use identical start--goal states, control limits, online RGB-D map updates, and A* replanning as the initially partial Gaussian map evolves. As shown in Fig.~\ref{fig:perception_comparison}, our method reaches the goal in $74$\,s, compared with $82$\,s for CAAP and $103$\,s without active perception, despite comparable path lengths ($16.6$, $16.9$, and $16.4$\,m). The improvement is instead reflected in a higher mean velocity ($0.212$\,m/s) while maintaining the largest mean safety barrier ($\bar h_s=0.724$). Moreover, the proposed trajectory-conditioned objective maintains a positive mean perception barrier ($\bar h_p=0.005$), whereas CAAP yields $\bar h_p=-0.275$. These results indicate that directing sensing toward uncertainty relevant to anticipated motion reduces conservative behavior without sacrificing safety.

\begin{figure}[t]
    \centering
    \includegraphics[
        width=\columnwidth,height=0.19\textheight,
        trim={0.25cm 0.30cm 0.20cm 0.20cm},
        clip
    ]{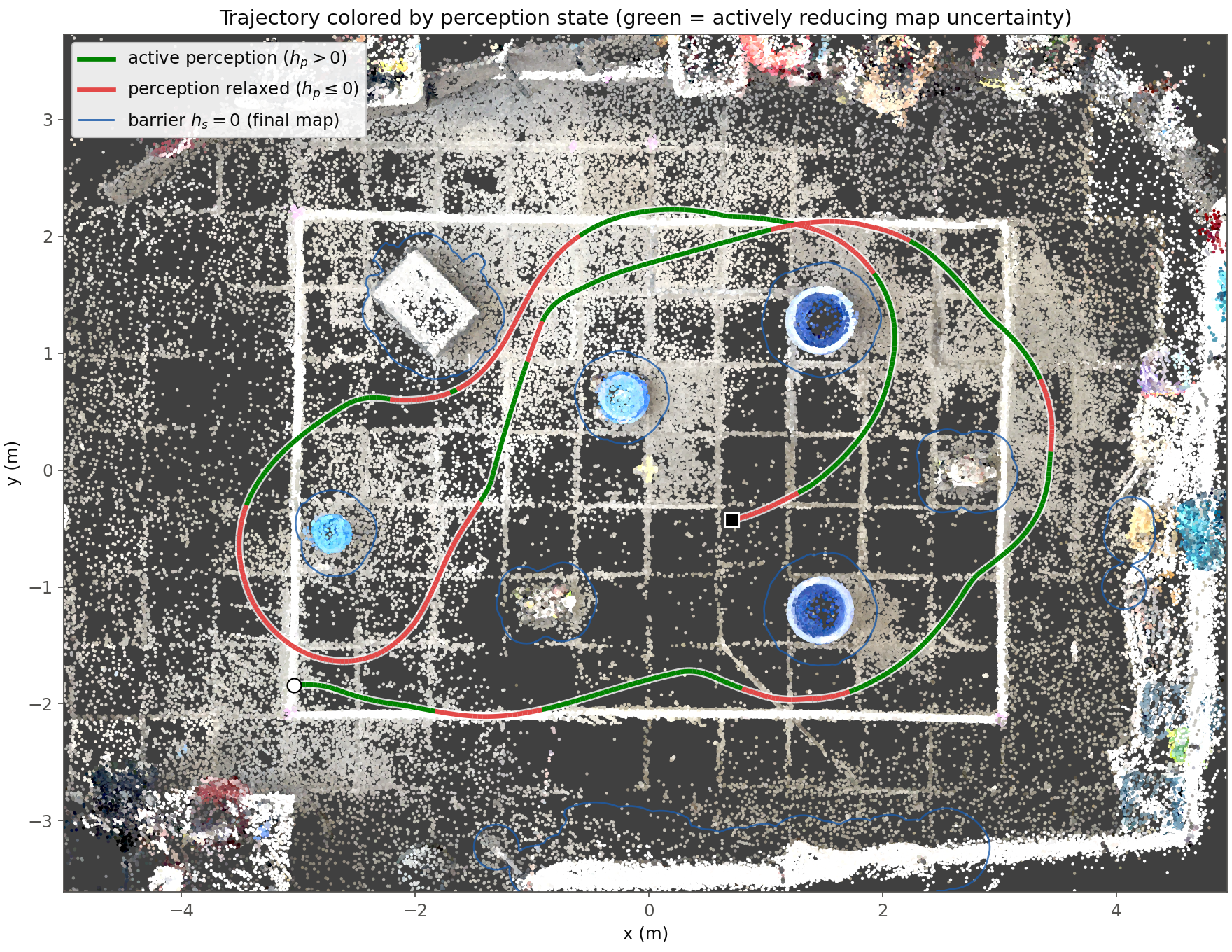}
    \vspace{-6.5mm}
\caption{
Real-robot Ackermann navigation in an indoor 3DGS map.
The executed trajectory is colored by the perception state: green denotes active perception ($h_p>0$) and red denotes perception relaxation ($h_p\leq0$). Blue contours indicate the safety boundary $h_s=0$. The controller regulates motion around unsafe regions while progressing from the start to the goal.
}
\label{fig:ackermann_trajectory}
    \vspace{-2mm}
\end{figure}

\begin{figure}[t]
    \centering
    \includegraphics[
        width=\columnwidth,height=0.13\textheight,
        trim={0.15cm 0.25cm 0.15cm 0.05cm},
        clip
    ]{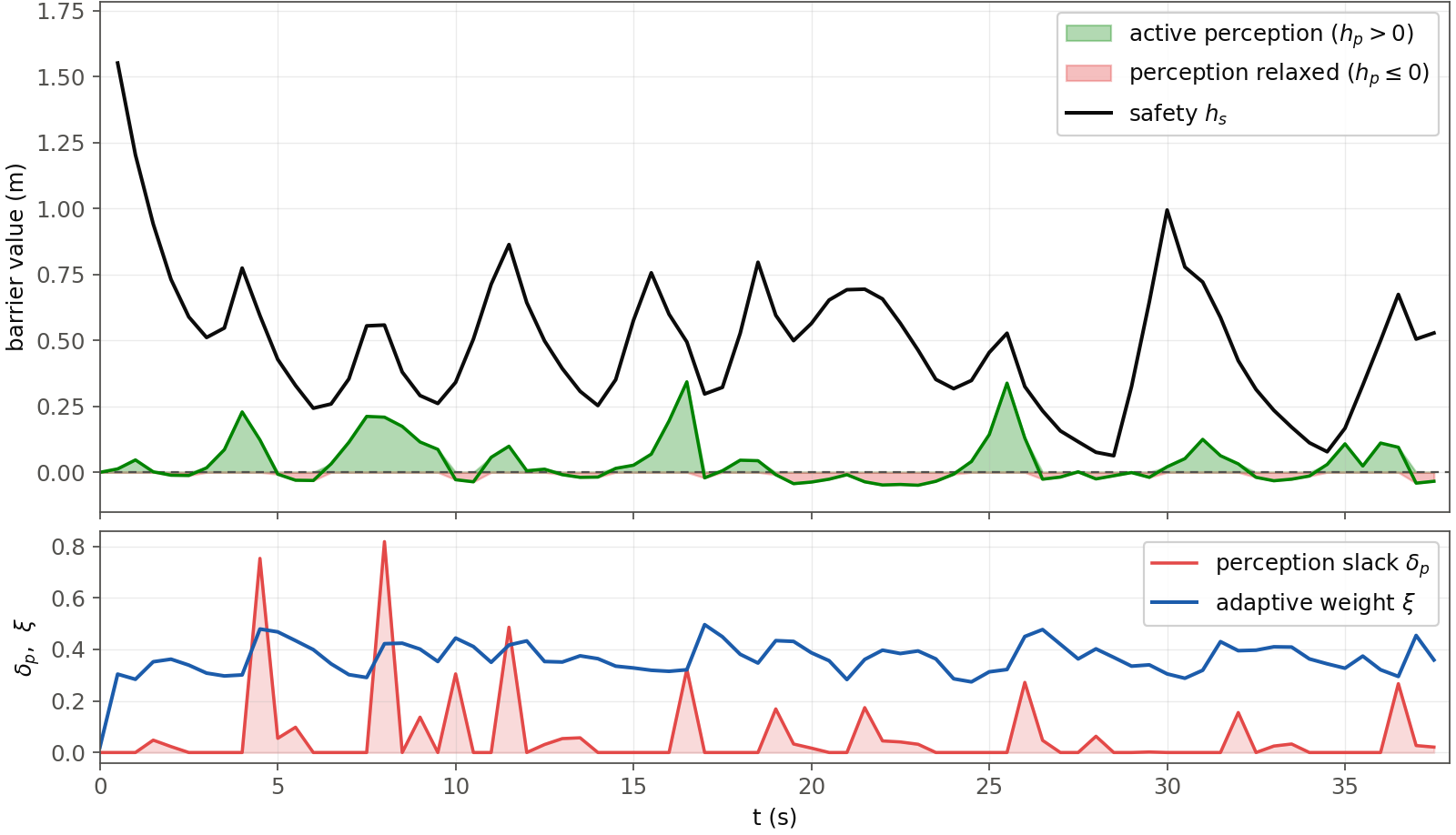}
    \vspace{-5.5mm}
\caption{
Barrier and perception-relaxation signals during the real-robot experiment.
The safety barrier $h_s$ remains positive throughout the run, while $h_p$, the perception slack $\delta_p$, and adaptive weight $\xi$ characterize the online safety--perception interaction.
}
\label{fig:ackermann_barrier}
    \vspace{-2mm}
\end{figure}

\subsection{Real-Robot Ackermann Validation}
Finally, we deploy the controller on a physical Ackermann steered mobile robot operating in an indoor 3DGS map. Unlike the unicycle simulation, the platform is subject to nonholonomic steering, bounded acceleration, and limited braking authority; the safety condition is therefore implemented through a higher order barrier while retaining the soft trajectory relevant perception constraint. Fig.~\ref{fig:ackermann_trajectory} shows that the controller departs from the nominal motion near safety-critical regions while continuing toward the goal. Over the $38$\,s run, the safety barrier remains strictly positive with a minimum value of $h_s=0.046$\,m; Fig.~\ref{fig:ackermann_barrier} further shows the interaction between safety enforcement and perception relaxation during execution. The hardware results demonstrate that the proposed map-level controller can be executed online under nonholonomic vehicle dynamics while maintaining the prescribed safety condition.

\subsection{Semantic aware safe active perception}
To incorporate semantics into safety-critical navigation, each Gaussian is augmented with a semantic attribute $\xi_i$. Using InteriorGS~\cite{InteriorGS2025}, we associate each primitive with the annotated object instance containing its center $\mu_i$ and assign the corresponding semantic class; unmatched primitives are labeled \emph{unknown}. Each semantic class $c$ is mapped to a positive risk-scaling factor $\kappa(c)$, with smaller values assigned to classes requiring more conservative treatment. The resulting semantic weighting modifies the relative contribution of each Gaussian to the composite safety barrier in~\eqref{eq:safety_barrier}. More generally, semantic attributes may be transferred from 3D annotations or lifted from image-level predictions~\cite{qin2024langsplat,wu2024opengaussian,yang2025opengs}; for datasets such as Gibson~\cite{Xia_2018_CVPR}, they can be aggregated from pixel-wise semantic observations across views. Thus, the controller is agnostic to the semantic estimation pipeline and
can incorporate semantic or VLM derived attributes as class dependent
collision-risk weights.~\cite{shorinwa2024splat,yu2025hammer,chen2025grad}.

We evaluate the effect of semantic collision risk by comparing the proposed
semantic-aware controller with the same safe-active perception policy using
geometry-only safety. As shown in Fig.~\ref{fig:semantic_risk_comparison}, semantic attributes modify the relative risk weighting of nearby Gaussian primitives and consequently alter the executed trajectory around semantically safety-critical objects. Quantitative differences in path length, goal time, and the mean and minimum safety barriers are reported in the figure.


\section{Conclusion}

We presented a semantic-risk-aware safe-active perception framework for navigation in attributed 3D Gaussian maps. Semantic attributes scale the risk-adjusted clearance of individual Gaussian primitives, allowing the composite safety CBF to prioritize obstacles according to both geometric uncertainty and semantic collision consequence. Experiments demonstrate navigation efficiency through active perception, semantic-dependent changes in the executed trajectory.


\newpage

\bibliographystyle{IEEEtran}
\bibliography{references}

@inproceedings{qin2024langsplat,
  title={Langsplat: 3d language gaussian splatting},
  author={Qin, Minghan and Li, Wanhua and Zhou, Jiawei and Wang, Haoqian and Pfister, Hanspeter},
  booktitle={Proceedings of the IEEE/CVF Conference on Computer Vision and Pattern Recognition},
  pages={20051--20060},
  year={2024}
}

@article{wu2024opengaussian,
  title={Opengaussian: Towards point-level 3d gaussian-based open vocabulary understanding},
  author={Wu, Yanmin and Meng, Jiarui and Li, Haijie and Wu, Chenming and Shi, Yahao and Cheng, Xinhua and Zhao, Chen and Feng, Haocheng and Ding, Errui and Wang, Jingdong and others},
  journal={Advances in Neural Information Processing Systems},
  volume={37},
  pages={19114--19138},
  year={2024}
}

@inproceedings{li2024gs3lam,
  title={Gs3lam: Gaussian semantic splatting slam},
  author={Li, Linfei and Zhang, Lin and Wang, Zhong and Shen, Ying},
  booktitle={Proceedings of the 32nd ACM International Conference on Multimedia},
  pages={3019--3027},
  year={2024}
}

@article{kerbl20233d,
  title={3d gaussian splatting for real-time radiance field rendering.},
  author={Kerbl, Bernhard and Kopanas, Georgios and Leimk{\"u}hler, Thomas and Drettakis, George and others},
  journal={ACM Trans. Graph.},
  volume={42},
  number={4},
  pages={139--1},
  year={2023}
}

@article{khass2026conflict,
  title={Conflict-Aware Active Perception and Control in 3D Gaussian Splatting Fields via Control Barrier Functions},
  author={Khass, Amirhossein Mollaei and Cosse, Athanasios and Pandey, Vivek and Motee, Nader},
  journal={arXiv preprint arXiv:2605.20566},
  year={2026}
}

@article{molnar2023composing,
  title={Composing control barrier functions for complex safety specifications},
  author={Molnar, Tamas G and Ames, Aaron D},
  journal={IEEE Control Systems Letters},
  volume={7},
  pages={3615--3620},
  year={2023},
  publisher={IEEE}
}

@article{khass2025active,
  title={Active Next-Best-View Optimization for Risk-Averse Path Planning},
  author={Khass, Amirhossein Mollaei and Liu, Guangyi and Pandey, Vivek and Jiang, Wen and Lei, Boshu and Daniilidis, Kostas and Motee, Nader},
  journal={arXiv preprint arXiv:2510.06481},
  year={2025}
}

@inproceedings{jiang2024fisherrf,
  title={Fisherrf: Active view selection and mapping with radiance fields using fisher information},
  author={Jiang, Wen and Lei, Boshu and Daniilidis, Kostas},
  booktitle={European Conference on Computer Vision},
  pages={422--440},
  year={2024},
  organization={Springer}
}

@article{chen2025splat,
  title={Splat-nav: Safe real-time robot navigation in gaussian splatting maps},
  author={Chen, Timothy and Shorinwa, Ola and Bruno, Joseph and Swann, Aiden and Yu, Javier and Zeng, Weijia and Nagami, Keiko and Dames, Philip and Schwager, Mac},
  journal={IEEE Transactions on Robotics},
  year={2025},
  publisher={IEEE}
}

@inproceedings{chen2025control,
  title={A Control Barrier Function for Safe Navigation with Online Gaussian Splatting Maps},
  author={Chen, Timothy and Swann, Aiden and Yu, Javier and Shorinwa, Ola and Murai, Riku and Kennedy, Monroe and Schwager, Mac},
  booktitle={2025 IEEE International Conference on Robotics and Automation (ICRA)},
  pages={11758--11765},
  year={2025},
  organization={IEEE}
}

@article{khass2026multi,
  title={Multi-Agent Next-Best-View Optimization for Risk-Averse Planning},
  author={Khass, Amirhossein Mollaei and Pandey, Vivek and Liu, Guangyi and Cosse, Athanasios and Bayrak, Emrah and Motee, Nader},
  journal={arXiv preprint arXiv:2606.04158},
  year={2026}
}

@article{Xu2024SplatfactoWAN,
  title={Splatfacto-W: A Nerfstudio Implementation of Gaussian Splatting for Unconstrained Photo Collections},
  author={Congrong Xu and Justin Kerr and Angjoo Kanazawa},
  journal={ArXiv},
  year={2024},
  volume={abs/2407.12306},
}

@misc{InteriorGS2025,
  title        = {InteriorGS: A 3D Gaussian Splatting Dataset of Semantically Labeled Indoor Scenes},
  author       = {SpatialVerse Research Team, Manycore Tech Inc.},
  year         = {2025},
  howpublished = {\url{https://huggingface.co/datasets/spatialverse/InteriorGS}}
}

@inproceedings{yang2025opengs,
  title={Opengs-slam: Open-set dense semantic slam with 3d gaussian splatting for object-level scene understanding},
  author={Yang, Dianyi and Gao, Yu and Wang, Xihan and Yue, Yufeng and Yang, Yi and Fu, Mengyin},
  booktitle={2025 IEEE International Conference on Robotics and Automation (ICRA)},
  pages={8486--8492},
  year={2025},
  organization={IEEE}
}

@InProceedings{Xia_2018_CVPR,
author = {Xia, Fei and Zamir, Amir R. and He, Zhiyang and Sax, Alexander and Malik, Jitendra and Savarese, Silvio},
title = {Gibson Env: Real-World Perception for Embodied Agents},
booktitle = {Proceedings of the IEEE Conference on Computer Vision and Pattern Recognition (CVPR)},
month = {June},
year = {2018}
}

@article{chen2025grad,
  title={Grad-nav++: Vision-language model enabled visual drone navigation with gaussian radiance fields and differentiable dynamics},
  author={Chen, Qianzhong and Gao, Naixiang and Huang, Suning and Low, JunEn and Chen, Timothy and Sun, Jiankai and Schwager, Mac},
  journal={IEEE Robotics and Automation Letters},
  volume={11},
  number={2},
  pages={1418--1425},
  year={2025},
  publisher={IEEE}
}

@article{yu2025hammer,
  title={Hammer: heterogeneous, multi-robot semantic gaussian splatting},
  author={Yu, Javier and Chen, Timothy and Schwager, Mac},
  journal={IEEE Robotics and Automation Letters},
  volume={10},
  number={7},
  pages={7270--7277},
  year={2025},
  publisher={IEEE}
}

@article{shorinwa2024splat,
  title={Splat-mover: Multi-stage, open-vocabulary robotic manipulation via editable gaussian splatting},
  author={Shorinwa, Ola and Tucker, Johnathan and Smith, Aliyah and Swann, Aiden and Chen, Timothy and Firoozi, Roya and Kennedy III, Monroe and Schwager, Mac},
  journal={arXiv preprint arXiv:2405.04378},
  year={2024}
}


\end{document}